\documentclass[conference]{IEEEtran}
\IEEEoverridecommandlockouts
\usepackage{cite}
\usepackage{amsmath,amssymb,amsfonts}
\usepackage{algorithmic}
\usepackage{graphicx}
\usepackage{textcomp}
\usepackage{xcolor}
\usepackage{hyperref}
\usepackage{booktabs}
\usepackage{pifont}
\usepackage{xcolor}
\usepackage{subfig}     % 子图包，不要与{subfigure}混用，{subfig}较新
\usepackage{overpic}     % 叭啦叭啦，与图片排版相
\usepackage{float}     %设置图片浮动位置的宏包
\usepackage{fontawesome}

\hypersetup{
    colorlinks=true,
    linkcolor=red,
    citecolor = red,
    filecolor=magenta,      
    urlcolor=cyan,
    }
\definecolor{red}{HTML}{c21313} % 
\newcommand{\cmark}{\textcolor{green!70!black}{\checkmark}}
\newcommand{\xmark}{\textcolor{red}{\ding{55}}}
\def\BibTeX{{\rm B\kern-.05em{\sc i\kern-.025em b}\kern-.08em
    T\kern-.1667em\lower.7ex\hbox{E}\kern-.125emX}}
\begin{document}

\title{Med-GLIP: Advancing Medical Language-Image Pre-training with Large-scale Grounded Dataset
}

% \author{\IEEEauthorblockN{\large anonymous authors}}
{
\author{\IEEEauthorblockN{Ziye Deng*}
\IEEEauthorblockA{\textit{ Zhejiang University}\\
Hangzhou, China \\
ziye.21@intl.zju.edu.cn}
\thanks{* These authors contributed equally to this work.}
\and
\IEEEauthorblockN{ Ruihan He*}
\IEEEauthorblockA{\textit{ Zhejiang University}\\
Hangzhou, China \\
ruihanhe021@gmail.com}
\and
\IEEEauthorblockN{ Jiaxiang Liu*}
\IEEEauthorblockA{\textit{ Guangdong Institute of Intelligence} \\
\textit{ Science and Technology}\\
Zhuhai, China \\
forworkliu@gmail.com}
\and
\IEEEauthorblockN{ Yuan Wang*}
\IEEEauthorblockA{\textit{ Zhejiang University}\\
 Hangzhou, China \\
 yuanwang23@zju.edu.cn}
\and 
\IEEEauthorblockN{ Zijie Meng}
\IEEEauthorblockA{\textit{ Zhejiang University}\\
 Hangzhou, China \\
 zijie.22@intl.zju.edu.cn}
\and
\IEEEauthorblockN{ Songtao Jiang}
\IEEEauthorblockA{\textit{ Zhejiang University}\\
 Hangzhou, China \\
 songtao.22@intl.zju.edu.cn}
\and
\IEEEauthorblockN{ Yong Xie \ \faEnvelope}
\IEEEauthorblockA{\textit{ Nanjing University of Posts} \\
\textit{ and Telecommunications}\\
 Nanjing, China \\
 yongxie@njupt.edu.cn}
\and
\IEEEauthorblockN{ Zuozhu Liu \ \faEnvelope}
\IEEEauthorblockA{\textit{Zhejiang University}\\
Hangzhou, China \\
zuozhuliu@intl.zju.edu.cn}
}}
\maketitle

\begin{abstract}
Medical image grounding aims to align natural language phrases with specific regions in medical images, serving as a foundational task for intelligent diagnosis, visual question answering (VQA), and automated report generation(MRG). However, existing research is constrained by limited modality coverage, coarse-grained annotations, and the absence of a unified, generalizable grounding framework. 
To address these challenges, we construct a large-scale medical grounding dataset~\textbf{Med-GLIP-5M} comprising over 5.3 million region-level annotations across seven imaging modalities, covering diverse anatomical structures and pathological findings. The dataset supports both segmentation and grounding tasks with hierarchical region labels, ranging from organ-level boundaries to fine-grained lesions.
Based on this foundation, we propose~\textbf{Med-GLIP}, a modality-aware grounding framework trained on Med-GLIP-5M.
%, and \textbf{Med-GLIP-5M}. 
Rather than relying on explicitly designed expert modules, Med-GLIP implicitly acquires hierarchical semantic understanding from diverse training data—enabling it to recognize multi-granularity structures, such as distinguishing lungs from pneumonia lesions. 
Extensive experiments demonstrate that Med-GLIP consistently outperforms state-of-the-art baselines across multiple grounding benchmarks. Furthermore, integrating its spatial outputs into downstream tasks, including medical VQA and report generation, leads to substantial performance gains. Our dataset is available at \href{https://www.modelscope.cn/datasets/Venn2025/Med-GLIP-5M}{Venn2025/Med-GLIP-5M}.
\end{abstract}


\section{Introduction}
With the rapid progress of medical AI, establishing precise alignment between natural language descriptions and specific regions in medical images has become a foundational step for tasks such as intelligent diagnosis, surgical navigation, and multimodal question answering~\cite{liu2021slake,zou2024medrg,chen2023medical,jiang2024moe,jiang2024med,jiang2025omniv,jiang2025hulu,jiang2019tiger,yeh2018unsupervised,chen2024r}. Medical image grounding aims to localize anatomical structures or pathological findings based on language input, enabling spatial-semantic correspondence across modalities. For example, when a radiologist states “a tumor is located in the upper right corner,” the model must accurately identify the corresponding image region. This task not only improves model interpretability but also provides critical spatial priors for downstream applications like visual question answering (VQA) and medical report generation (MRG), playing an essential role in clinical decision support\cite{wang2025fair,chen2024r,jiang2024joint,gai2025medthink,liu2024medcot,liu2024vpl,liu2023parameter,liu2025kpl,zhang2025medu1incentivizingunifiedmedical}

%zijie
However, compared to the natural image domain, medical image grounding faces several unique challenges~\cite{sun2024medical,khan2025comprehensive}. First, publicly available grounding datasets are extremely scarce, especially those with large-scale, multi-organ, and multi-modality annotations, which significantly limit progress in this field. Second, medical images are highly specialized and exhibit substantial heterogeneity across modalities such as CT, MRI, and ultrasound—in terms of spatial resolution, anatomical appearance, and contrast—making cross modal alignment inherently complex~\cite{tong2023tsnet,liu2023deep}. Moreover, many target regions, such as small lesions or vascular branches, often lack clear boundaries and exhibit high anatomical variability, placing greater demands on the model’s precision and generalization capabilities.

\begin{table*}[t]
\centering
\caption{Unified comparison of medical image \emph{grounding} datasets. BB = bounding-box, 3D-BB = volumetric BB, EP = extreme-point supervision, ROI = region-of-interest. \cmark/\xmark: support / not supported.}
\label{tab:grounding-unified}
\resizebox{\linewidth}{!}{
\begin{tabular}{c c c c c c c c p{5.5cm}} 
\toprule
\textbf{Dataset} & \textbf{Year} & \textbf{RoI Scale} & \textbf{Annotation Type} & \textbf{Multi-modal} & \textbf{Seg.} & \textbf{Ground} & \textbf{$\ge$100K} & \textbf{Coverage / Highlight} \\ 
\midrule
VQA-RAD          & 2018 & N/A     & None        & \xmark & \xmark & \xmark & \xmark & Common pathologies; QA \\
SLAKE            & 2021 & 642 images     & Mask        & \xmark & \cmark & \xmark & \xmark & 7 organ categories; QA + segmentation \\
MS-CXR           & 2022 & 1162 images    & BB          & \xmark & \xmark & \cmark & \xmark & 8 thoracic findings \\
M3D-Seg           & 2023 & 10410 studies  & 3D-BB       & \xmark & \xmark & \cmark & \xmark & Volumetric abnormalities \\
MedTrinity-25M   & 2024 & –              & BB / Mask   & \cmark & \cmark & \cmark & \xmark & $>$15 organs, multimodal reports \\
\textbf{Med-GLIP-5M} & 2025 & 5.3M pairs     & BB / Mask   & \cmark & \cmark & \cmark & \cmark & 7 modalities, 30+ anatomical regions \\
\bottomrule
\end{tabular}
}
\end{table*}

To address data scarcity and semantic misalignment in medical image grounding, prior works have explored both dataset construction and cross-modal modeling. On the dataset side, SLAKE~\cite{liu2021slake} offers region-phrase annotations for chest X-rays, marking an early attempt at grounding, but it contains only 6k samples and has limited modality coverage. MedTrinity-25M~\cite{xie2024medtrinity} provides large-scale but loosely aligned image-text pairs across tasks, lacking fine-grained region-level supervision. These datasets commonly suffer from: (1) limited modality and organ diversity; (2) absence of dense region-level annotations; and (3) a focus on classification or QA tasks rather than spatial-semantic alignment. On the modeling side, recent approaches adapt natural-image cross-modal models to the medical domain (e.g., LLaVA-Med~\cite{li2023llavamed}, MedKLIP~\cite{wu2023medklip}, MedSAM~\cite{medsam}), often incorporating structured medical knowledge. LLaVA-Med~\cite{li2023llavamed} introduces multimodal LLMs but relies on image-caption pairs without fine-grained alignment. MedKLIP~\cite{wu2023medklip} employs region-phrase contrastive learning but remains X-ray–specific. MedSAM~\cite{medsam} incorporates structure-aware priors for zero-shot organ segmentation based on SAM~\cite{sam}, yet it depends on prompts and lacks deep language grounding. OntoRay further integrates radiological ontologies to capture causal relations between terms and regions~\cite{chepelev2023ontologies}. Despite these efforts, a unified, modality-adaptive grounding framework capable of multi-scale semantic alignment and inter-modality generalization is still missing, and its impact on downstream tasks such as VQA or report generation remains underexplored.

In summary, our contributions are as follows:

\begin{figure}[t]
  \centering
  \includegraphics[width=0.9\linewidth]{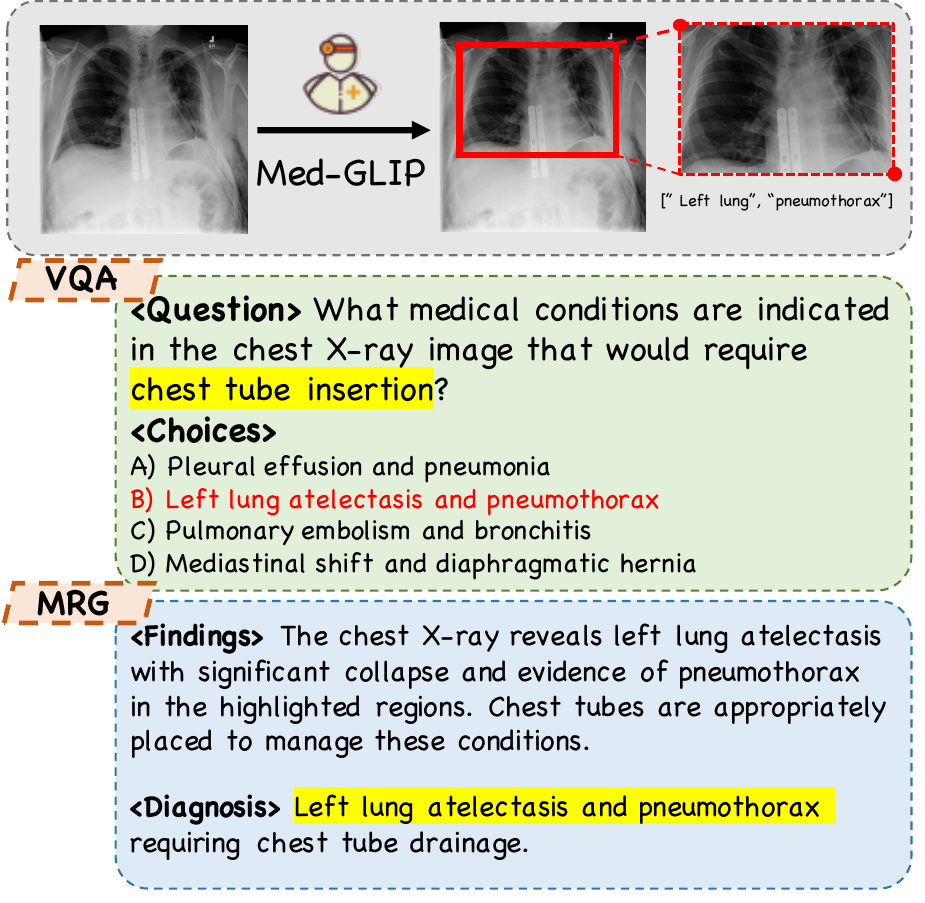}
  \caption{Enhance VQA and MRG with Med-GLIP.}
\label{teafigure}
\end{figure}

\begin{itemize}
	\item 	To the best of our knowledge, we introduce the largest and most diverse medical grounding dataset to date, \textbf{Med-GLIP-5M}, spanning 7 modalities, 30+ anatomical regions, and over 5.3 million image-text region pairs, addressing the long-standing data scarcity in this field (\autoref{tab:grounding-unified}).
	\item 	We propose \textbf{Med-GLIP}, a modality-aware hierarchical expert framework that enables high-precision grounding across diverse medical imaging types by modeling fine-grained structural differences (\autoref{figure3}).
	\item 	Through extensive experiments, we validate the effectiveness of spatial grounding in downstream tasks, showing that Med-GLIP enhances both VQA and report generation performance (\autoref{teafigure}). Our work provides a unified solution across data, modeling, and application perspectives, advancing the development of generalizable medical vision-language models.
\end{itemize}

\section{Med-GLIP-5M Construction}

\begin{figure*}[t!]
  \centering
  \includegraphics[width=\linewidth]{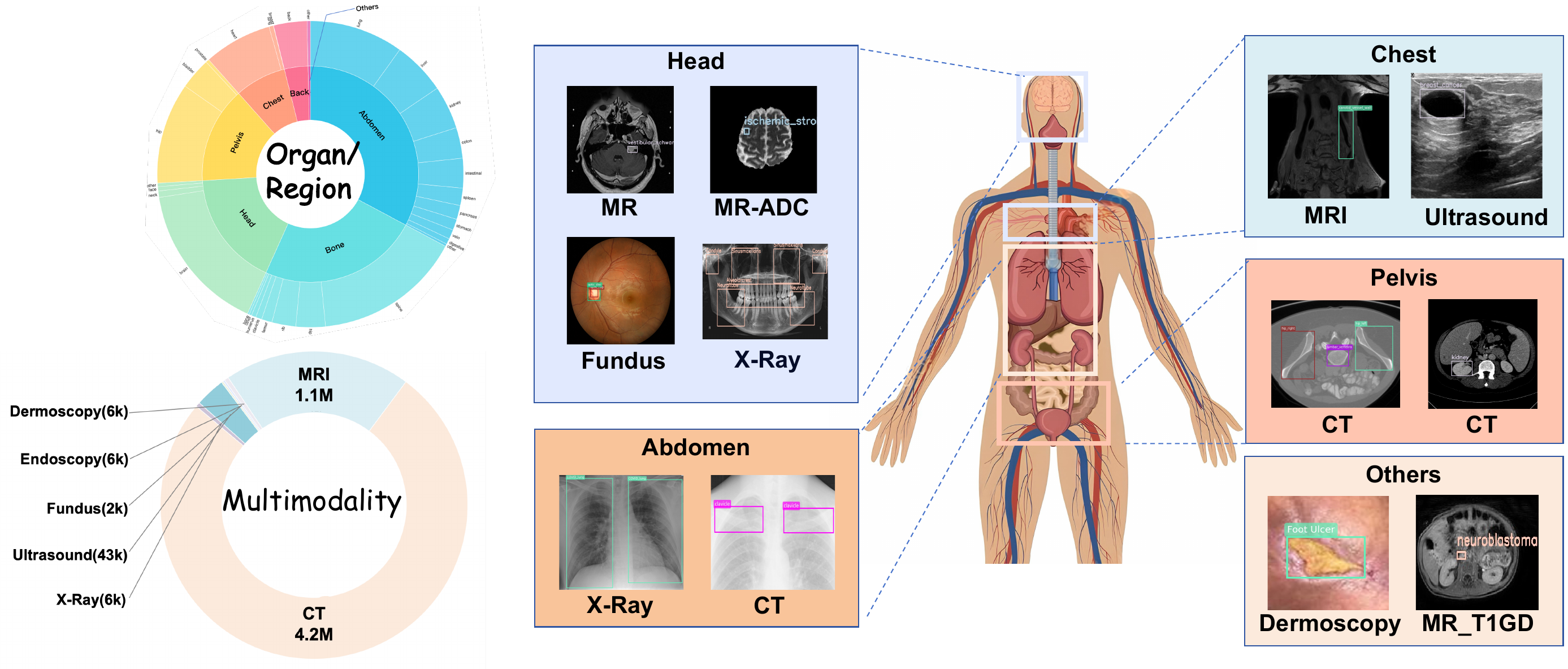}
  \caption{Med-GLIP-5M has 7 modality categories in total, with multiple organs containing suborgans in the dataset.}
  \label{figure2}
\end{figure*}

\subsection{Data Collection}
We curated a multi-source medical imaging dataset for vision-language grounding. Each image was selected with a preference for multi-instance spatial annotations—either multiple masks or bounding boxes per image—thus ensuring the resulting dataset is well-suited for vision-language grounding tasks. Datasets were extracted from several platforms including but not limited to: GitHub, Grand Challenge , OpenNeuro, BraTS, SYNAPSE, and CodaLab. Selection criteria: (1) DICOM/NIfTI accessibility; (2) spatial annotations (masks/bounding boxes); (3) coverage of multiple anatomical regions. This yielded 2720979 images spanning X-ray, CT, MRI across multiple organs including abdomen(32.8\%), bone(23.9\%), head(17.3\%), chest(7.6\%) etc. The final dataset has 4.13 masks/image on average, with various modality including CT(61.84\%), MRI(26.71\%), Ultrasound(1.44\%), X-RAY(0.22\%), Dermoscopy(0.21\%), Endoscopy(0.09\%), Fundus(0.09\%). The collected datasets span a wide range of imaging modalities, spatial resolutions, and anatomical regions, ensuring diversity across organs and clinical tasks, as shown in \autoref{figure2}.

\subsection{Data Quality Control}
A three-tier quality control (QC) pipeline was implemented post-collection. Firstly, we discarded non-readable or broken files, sliced 3D datasets into 2D to improve data quantity, and standardized all image formats to PNG while preserving the original resolution. Secondly, we verified that each image had a corresponding and aligned segmentation mask, excluding samples that failed this check. Thirdly, we filtered inconsistent annotations, such as those with malformed structures or undefined semantics. Moreover, we removed samples where the annotated mask covered only a negligible portion of the image (empirically set as 1.5\% area exclusion), as such cases do not provide meaningful spatial grounding for downstream tasks\cite{Everingham15}. This quality assurance workflow ensures that only well-structured, clearly labeled, and spatially meaningful samples are retained for model training and evaluation.

\subsection{Data Preprocessing}

\begin{figure*}[ht!]
  \centering
  \includegraphics[width=\linewidth]{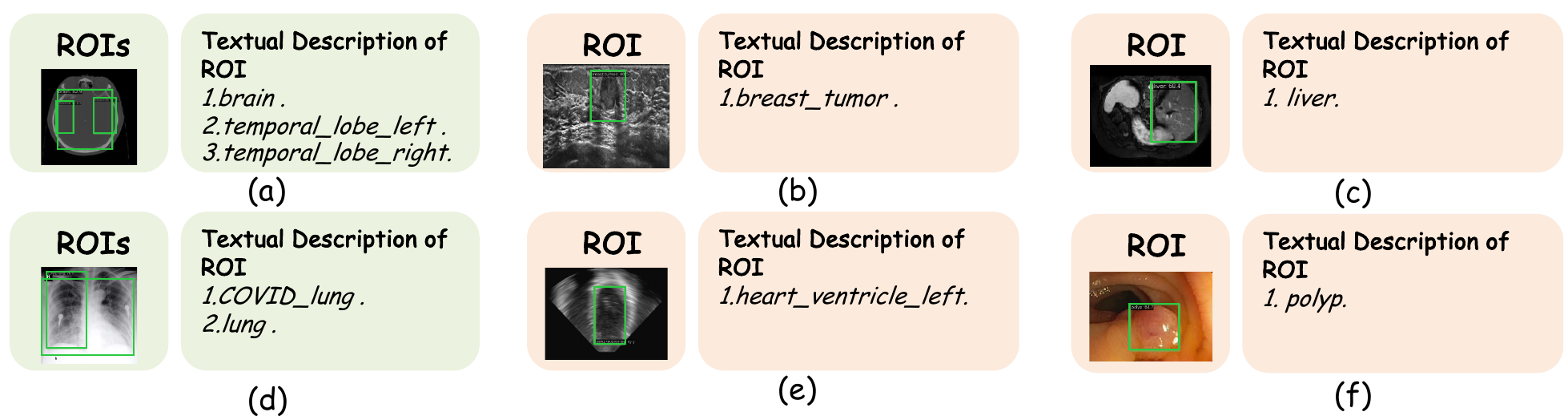}
  \caption{Illustration of hierarchical region-level annotations across modalities.
Each subfigure (a–f) shows green bounding boxes and textual descriptions over CT, X-ray, ultrasound, endoscopy, and MRI images. Multi-level boxes reflect hierarchical semantics, providing fine-grained region-text supervision for structured medical grounding.}
\label{figure3}
\end{figure*}

The data preprocessing phase for Med-GLIP-5M was designed to ensure high-quality data while adhering to ethical standards and regulatory requirements. After acquisition, datasets underwent a multi-stage refinement process. Initially, data were categorized by organ type and volume, with duplicates and low-quality samples removed. Semantic label inconsistencies were resolved through standardized naming. Clean data were then augmented with geometric and intensity transformations to enhance diversity without compromising diagnostic integrity. Image formats were standardized to ensure compatibility across machine learning frameworks. A rigorous validation protocol verified the consistency between images and segmentation masks. Additionally, clinical metadata such as patient history and diagnostic outcomes were integrated where available. This structured approach resulted in a dataset that is both large-scale and high-quality, as visualized in \autoref{figure3}, providing a robust foundation for medical AI research.

Finally, all processed datasets were merged and reformatted into a unified metadata repository conforming to the COCO standard. This consolidated dataset serves as the foundation for training our Med-GLIP model, enabling robust and scalable medical vision-language learning across heterogeneous clinical data sources.

\subsection{Data Statistics}

Med-GLIP-5M comprises 198 fine-grained annotation labels, which were re-organized into 38 broader anatomical categories. These 38 hierarchical classes span 6 major body regions, with a cumulative image count exceeding 11 million. This hierarchical label structure allows for flexible experimentation, such as organ-level segmentation and multi-organ detection. The dataset exhibits substantial heterogeneity with 7 distinct imaging modalities. Computed Tomography (CT) and Magnetic Resonance Imaging (MRI) are predominant, contributing approximately 4.2 million and 1.08 million images, respectively, and collectively account for over 80\% of the total dataset volume. The remaining modalities, such as ultrasound, contribute important diversity.

This multimodal composition makes the dataset highly suitable for a broad spectrum of downstream applications, including cross-modal learning, domain adaptation, modality-aware segmentation, and multimodal fusion in disease detection and prognosis. It also enables research into unified frameworks that bridge traditionally distinct clinical domains.

\section{Med-GLIP}
\label{section:method}

In our work, medical object detection is reconceived as a phrase grounding task, wherein each image region that is identified corresponds to its matching medical phrase. Given a predefined set of medical concepts pertinent to a particular imaging modality, for instance, \{``pneumonia'', ``nodule'', ``fracture''\} for X-ray images, a prompt is constructed: 
$$\text{Prompt} = \text{``Detect: pneumonia, nodule, fracture''}. $$

Following methodologies similar to GLIP \cite{li2022grounded}, we can employ pre-trained language models, e.g., BERT \cite{devlin2018bert}, to encode more semantically rich prompts (e.g., ``pneumonia. nodule. fracture.''), which has demonstrated empirical advantages. Within our modality-specific grounding framework, alignment scores $S_{\text{ground}}$
are computed between medical image region features $\mathbf{F}$ and the encoded word or token features $\mathbf{T}$ from the prompt. This is formally expressed as:
\begin{equation}
\label{eqn:ground_logits_revised}
\mathbf{F} = \text{Enc}_{I}(\text{Img}),  \mathbf{T} = \text{Enc}_{L}(\text{Prompt}), S_{\text{ground}} = \sigma(\mathbf{F} \mathbf{T}^{\top})
\end{equation}

where $\mathbf{F} \in \mathbb{R}^{N \times d}$
 represents the region features extracted by an image encoder $\text{Enc}_{I}$
, and $\mathbf{T} \in \mathbb{R}^{M \times d}$
 denotes the contextual word/token features from a language encoder $\text{Enc}_{L}$
. $\sigma(\cdot)$ represents the sigmoid function for normalization. Each imaging modality (e.g., X-ray, CT, MRI) utilizes a dedicated image encoder $\text{Enc}_{I, \text{modality}}$
, while a common language encoder $\text{Enc}_{L}$
 is shared across modalities. The grounding model for each modality is trained end-to-end by minimizing the loss function defined in Equation \eqref{eqn:loss}, adapted such that the classification logits $S_{\text{cls}}$
 from Equation \eqref{eqn:cls_logits} are substituted with the computed alignment scores $S_{\text{ground}}$.
\begin{equation}\label{eqn:loss}
    L_{cal} = L_{cal}^{\text{cls}} + L_{cal}^{\text{loc}}.
\end{equation}
\begin{equation}\label{eqn:cls_logits}
    \mathbf{F} = \text{Enc}_{I}(\text{Img}), S_{\text{cls}} = \mathbf{F} W^T, L_{cal}^{\text{cls}} = loss(S_{\text{cls}}; T),
\end{equation}

Equation \eqref{eqn:loss} and \eqref{eqn:cls_logits} outline the loss computation for a baseline detection model. The model extracts features $\mathbf{F}$ via the image encoder $\mathrm{Enc}_I$ and computes classification logits $S_{cls}$ using a classifier weight matrix $\mathbf{W}$. The total training objective, $L_{cal}$, is a composite loss comprising two components: (i) a classification loss, $L_{cal}^{cls}$, calculated from the logits $S_{cls}$ and ground-truth targets $\mathbf{T}$, and (ii) a localization loss, $L_{cal}^{loc}$, used for bounding box regression.

To address the potential mismatch between the number of tokens ($M$) in the encoded prompt and the number of target medical concepts ($c$), we adopt an expansion strategy analogous to that in GLIP. For a binary sigmoid loss, the ground truth target matrix $\mathbf{T} \in \{0,1\}^{N \times c}$
 is expanded into $\mathbf{T}' \in \{0,1\}^{N \times M}$
. This expansion involves aligning all sub-word tokens constituting a positive medical phrase to be positive matches (i.e., target value of 1), while any additionally introduced tokens (e.g., punctuation, padding) are designated as negative matches (i.e., target value of 0). During the inference phase, the probability for each medical phrase is derived by aggregating the probabilities of its constituent tokens, typically by averaging. This allows for a flexible and robust mapping from textual prompts to visual detections.

\subsection{Modality-Specific Equivalence.} Within each medical imaging modality, our reformulation establishes a theoretical equivalence between object detection and phrase grounding. By training a grounding model on modality-specific data and prompts, we can directly apply it to detection tasks within that modality in a zero-shot manner, leveraging the rich semantic information encoded in the language prompt.

\section{Experiments}
\label{experiments}
% \input{paragraphs/tabs/tab_low}
% \input{paragraphs/algorithm}

% In this section, we present a comprehensive set of experiments and analyses to showcase the effectiveness of our proposed \Method \ scoring strategy in diverse scenarios. We begin by empirically evaluating our \method method against other baselines (\cref{sec:baseline}). Subsequently, we conduct a series of ablation studies to investigate the effectiveness of the proposed two main components: variance measurement and dual-window strategy (\cref{sec:ablation}). Additionally, we perform cross-architecture experiments to evaluate the robustness of our coresets, assessing their performance when selected on one architecture and tested on others.
% \input{paragraphs/tabs/tab_all_SA}
\subsection{Experiment Setup}
\label{subsec:experiment_setting}

\textbf{Datasets for Region-Level Grounding.} \quad
To evaluate the grounding performance of our GLIP-experts, we selected six datasets spanning five imaging modalities: SegRap2023\_ct (CT), AMOS2022\_MR (MRI), sz\_cxr (X-ray), CETUS2014 (Ultrasound), and CVC-ClinicDB (Endoscopy). All datasets except CVC-ClinicDB are sourced from the IMed-361M benchmark\cite{cheng2024interactive}, which provides standardized train-test splits (90\%/10\%). For evaluation, we use only the held-out test sets. CVC-ClinicDB lacks predefined splits and is used in its entirety. These datasets collectively cover 5,551 images with rich structural annotations across modalities.

\textbf{Datasets for MRG.} \quad
We adopt two widely-used benchmarks: MIMIC-CXR\cite{johnson2019mimic} and IU-Xray\cite{huang2023kiut}. MIMIC-CXR includes 377,110 chest X-ray images and corresponding free-text radiology reports from 227,835 studies. It is evaluated using the MLRG framework to compute both natural language generation (NLG) metrics and concept grounding scores (e.g., CE). IU-Xray, with 7,470 image-report pairs, offers a smaller and more structured alternative, and is evaluated using the R2Gen model on standard NLG metrics.
\begin{table*}[t]
    \renewcommand{\arraystretch}{1.1}
    \caption{Performance of different models on grounding tasks across multiple medical imaging modalities. Metrics are Average Precision (AP) and AP@50. "Zero" indicates zero-shot inference; Med-GLIP models are evaluated at 10\% and 100\% training scale.}
    \vspace{-5pt}
    \centering
    \setlength{\tabcolsep}{3mm}
    \begin{normalsize}
    \resizebox{\linewidth}{!}{
    \begin{tabular}{c|cc|cc|cc|cc|cc|cc|cc}
    \toprule[0.8pt]
    \textbf{Modality} & \multicolumn{2}{c|}{\textbf{CT}} & \multicolumn{2}{c|}{\textbf{MRI}} & \multicolumn{2}{c|}{\textbf{X-ray}} & \multicolumn{2}{c|}{\textbf{Ultrasound}} & \multicolumn{2}{c|}{\textbf{Endoscopy}} & \multicolumn{2}{c|}{\textbf{Dermoscopy}} & \multicolumn{2}{c}{\textbf{Fundus Photography}} \\
    \midrule
    \textbf{Representative Dataset} & \multicolumn{2}{c|}{SegRap} & \multicolumn{2}{c|}{AMOS2022} & \multicolumn{2}{c|}{Sz\_cxr} & \multicolumn{2}{c|}{CETUS2014} & \multicolumn{2}{c|}{Clinic-DB} & \multicolumn{2}{c|}{Rimonedl} & \multicolumn{2}{c}{isic2017} \\
    \textbf{Metric} & AP & AP50 & AP & AP50 & AP & AP50 & AP & AP50 & AP & AP50 & AP & AP50 & AP & AP50 \\
    \midrule
    GLIP-zero & 0.0 & 0.0 & 0.0 & 0.0 & 0.0 & 0.0 & 0.0 & 0.0 & 0.0 & 0.0 & 0.0 & 0.0 & 0.0 & 0.0 \\
    CO\_DERT-zero & 0.0 & 0.0 & 0.0 & 0.0 & 0.0 & 0.0 & 0.0 & 0.0 & 0.0 & 0.0 & 0.0 & 0.0 & 0.0 & 0.0 \\
    CO\_DERT\_100\% & 31.8 & 38.9 & 7.6 & 11.3 & 69.8 & 84.2 & 1.0 & 7.0 & \textbf{60.4} & \textbf{79.7} & 43.0 & 79.5 & \textbf{71.0} & \textbf{87.3} \\
    Med-GLIP\_10\% & 41.2 & 80.7 & 16.0 & 43.9 & 4.9 & 38.4 & 9.5 & 23.5 & 10.3 & 33.4 & 36.0 & 75.9 & 14.7 & 71.3 \\
    Med-GLIP\_100\% & \textbf{82.7} & \textbf{99.0} & \textbf{28.0} & \textbf{59.2} & \textbf{86.9} & \textbf{100.0} & \textbf{59.2} & \textbf{91.9} & \underline{33.4} & \underline{62.5} & \textbf{61.7} & \textbf{90.1} & \underline{58.8} & \underline{78.3} \\
    \bottomrule[0.8pt]
    \end{tabular}
    }
    \end{normalsize}
    \label{table2}
    \end{table*}

% \begin{figure*}[ht!]
%     \centering % 整体居中

%     % --- 左图 (Figure 4) ---
%     \begin{minipage}[b]{0.5\textwidth}
%         \centering
%         \includegraphics[width=\linewidth]{paragraphs/figs/MRG.pdf}
%         \caption{Performance comparison between w/ and w/o Med-GLIP in MRG Task on the baseline R2Gen and MLRG.}
%         \label{MRG Comparison}
%     \end{minipage}
%     \hfill % 在两个图之间添加弹性的水平间距
%     % --- 右图 (Figure 5) ---
%     \begin{minipage}[b]{0.48\textwidth}
%         \centering
%         \includegraphics[width=\linewidth]{paragraphs/figs/VQA.pdf}
%         \caption{Performance comparison between w/ and w/o Med-GLIP in Med-VQA on the VQA-RAD, SLAKE, PathVQA Dataset.}
%         \label{VQA_Comparison}
%     \end{minipage}
    
% \end{figure*}
 
\textbf{Datasets for Med-VQA.} \quad
We evaluate model generalization to vision-language reasoning via three medical VQA datasets: VQA-RAD\cite{lau2018dataset}, SLAKE\cite{liu2021slake}, and PathVQA\cite{he2020pathvqa}. These datasets span radiological and pathological domains, include both English and Chinese annotations, and provide over 50,000 QA pairs. During evaluation, GLIP-experts support both: (1) \textit{Closed-ended} multiple-choice reasoning by verifying spatial alignment between phrases and detected regions; and (2) \textit{Open-ended} answer generation through grounded semantic reasoning. All evaluations follow official train-test splits for reproducibility and fair comparison.

    \vspace{5pt}
    \textbf{Baselines and Models.} \ \ To evaluate the effectiveness of our medical bounding box dataset for downstream radiology report generation tasks, we conducted experiments using two representative models: R2Gen and MLRG. R2Gen\cite{chen2015generating} is a Transformer-based model for medical report generation. MLRG\cite{liu2025enhanced} is a recent state-of-the-art model that leverages multi-view longitudinal data and contrastive learning. 
    
    For Med-VQA tasks, we employed LLaVA-Med\cite{li2023llavamedtraininglargelanguageandvision}, a leading large language and vision assistant designed specifically for biomedical applications.
    
    For the grounding tasks, we compared 5 different models: (1) the original GLIP \cite{li2022grounded} without fine-tuning, (2) GLIP experts fine-tuned on modality-specific subsets, (3) GLIP experts fine-tuned on 10\% modality-specific subsets, (4) the original CO\_DETR model\cite{zong2023detrs} without fine-tuning, and (5) CO\_DETR model fine-tuned on modality-specific subsets.

% \begin{figure}[t]
% \centering\includegraphics[width=0.4\textwidth]{paragraphs/figs/med_glip_comparison.pdf}
% \caption{Performance comparison between GLIP and Med-GLIP on three medical image datasets (ClinicDB, ETIS, and Kvasir) in terms of AP@50. Med-GLIP significantly outperforms the original GLIP model across all datasets, demonstrating the effectiveness of domain adaptation for medical visual grounding tasks.} 
% \label{Med-Glip_Datasets}
% \end{figure}

% \begin{figure*}[ht!]
%     \centering % 整体居中

%     % --- 左图 (Figure 4) ---
%     \begin{minipage}[b]{0.48\textwidth}
%         \centering
%         \includegraphics[width=\linewidth]{paragraphs/figs/MRG.pdf}
%         \caption{Performance comparison between w/ and w/o Med-GLIP in MRG Task on the baseline R2Gen and MLRG.}
%         \label{MRG Comparison}
%     \end{minipage}
%     \hfill % 在两个图之间添加弹性的水平间距
%     % --- 右图 (Figure 5) ---
%     \begin{minipage}[b]{0.48\textwidth}
%         \centering
%         \includegraphics[width=\linewidth]{paragraphs/figs/VQA.pdf}
%         \caption{Performance comparison between w/ and w/o Med-GLIP in Med-VQA on the VQA-RAD, SLAKE, PathVQA Dataset.}
%         \label{VQA_Comparison}
%     \end{minipage}
    
% \end{figure*}

\textbf{Implementation Details.} \ \ 
All experiments were conducted on an Ubuntu server equipped with 8 NVIDIA RTX 3090 GPUs (24 GB each). 
For grounding tasks, GLIP and Co-DETR were fine-tuned on the Med-GLIP dataset for 30 epochs using the Adam optimizer with a learning rate of $2 \times 10^{-4}$ and a batch size of 5.
For medical report generation, R2Gen was trained for 100 epochs using StepLR scheduling, with a learning rate of $5 \times 10^{-5}$ for the vision encoder and $1 \times 10^{-4}$ for other parameters (batch size = 32). MLRG was trained for 50 epochs using AdamW and ReduceLROnPlateau, with a learning rate of $5 \times 10^{-5}$, and a batch size of 6.

    \vspace{5pt}
    \textbf{Evaluation Metric.} \ \ 
    To evaluate the precision of the grounding model GLIP, we adopted a metric called Average Precision (AP). It measures the area under the precision-recall curve. A higher AP indicates better overall detection or localization performance.

    We adopted 2 metrics to evaluate the quality of MRG results: the natural language generation (NLG) metric and the clinical efficacy (CE) metric. The NLG metric quantifies how closely a generated report matches the reference text in terms of linguistic similarity. It includes BLEU-n, METEOR, and ROUGE-L. The CE metric focuses on medical accuracy rather than linguistic overlap. It includes RadGraph F1 score, CheXpert F1 score, Precision (P), and Recall (R). For Med-VQA tasks, We employ the accuracy for closed-set questions and recall for open-set questions, being consistent with existing work like LLaVA-Med\cite{li2023llavamedtraininglargelanguageandvision} for a fair comparison.

    \subsection{Results \& Analysis}\label{subsec:experiment_setting}

    \vspace{5pt}
    \textbf{Grounding Performance across Models.} \ \  
    As shown in \autoref{table2}, we first compare the zero-shot grounding performance of GLIP and CO-DETR across all modalities and datasets. Both models achieve near-zero accuracy in this setting, indicating that neither is able to effectively localize medical entities without domain-specific fine-tuning. This demonstrates the substantial domain gap and highlights the necessity of fine-tuning for grounding tasks in medical imaging.
    
    When we finetune Med-GLIP with 10\% and 100\% of the training data, we observe a clear and consistent improvement in both AP and AP50 metrics across all modalities. For instance, on the CT (SegRap) dataset, AP increases from 0.0 (zero-shot) to 41.2 at 10\% scale, and further to 82.7 at 100\% scale. Similar trends are observed in X-ray (Sz\_cxr), Ultrasound (CETUS2014), endoscopy (Clinic-DB), and other modalities, confirming the effectiveness of our dataset and training strategy in progressively enhancing model performance as more annotated data becomes available.

After full finetuning, Med-GLIP outperforms CO-DETR in five out of seven modalities, including CT, MRI, X-ray, ultrasound, and dermoscopy. For instance, Med-GLIP achieves substantially higher AP scores on CT (82.7 vs. 31.8), MRI (28.0 vs. 7.6), X-ray (86.9 vs. 69.8), ultrasound (59.2 vs. 1.0), and dermoscopy (61.7 vs. 43.0). Although CO-DETR attains slightly better performance in endoscopy and fundus photography, Med-GLIP demonstrates robust and stable performance across a wider range of modalities, highlighting its strong generalizability and effectiveness as a unified framework for medical grounding tasks. Overall, these results demonstrate that our dataset and adaptation strategy benefit a variety of grounding models, while Med-GLIP in particular achieves the best performance (\autoref{table2}).

\begin{figure*}[ht!]
\centering\includegraphics[width=1\textwidth]{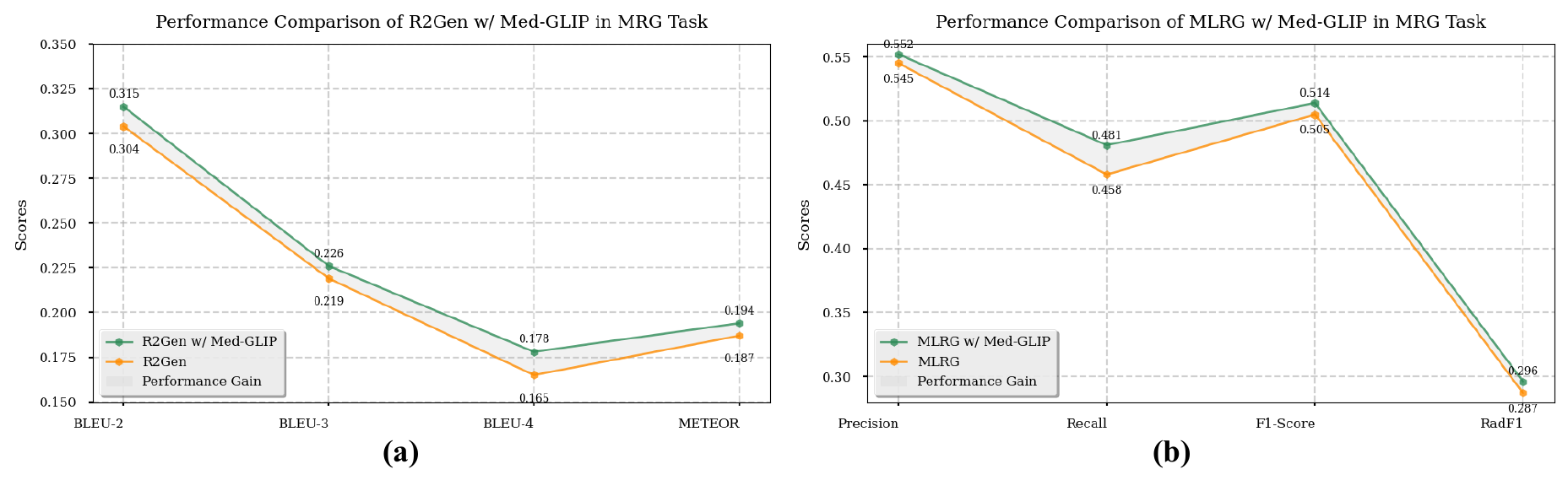}
\caption{Performance comparison between w/ and w/o Med-GLIP in MRG Task on the baseline R2Gen and MLRG.} 
\label{MRG Comparison}
\end{figure*}   

\begin{figure*}[ht!]
\centering
\includegraphics[width=0.9\textwidth]{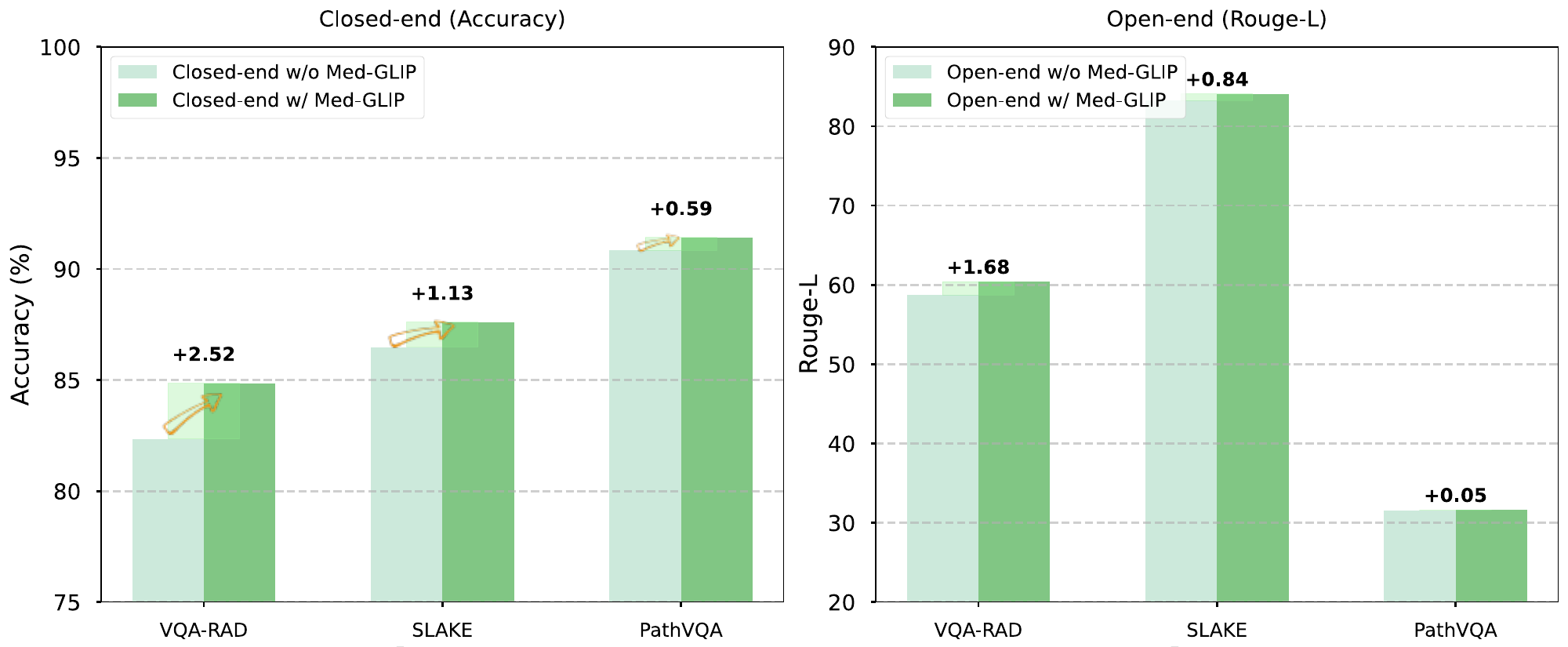}
\caption{Performance comparison between w/ and w/o Med-GLIP in Med-VQA on the VQA-RAD, SLAKE, PathVQA Dataset.} 
\label{VQA_Comparison}
\end{figure*}

    \vspace{5pt}
    \textbf{Performance of downstream MRG tasks} 
    As shown in \autoref{MRG Comparison}, the enhanced R2Gen model achieved higher BLEU and METEOR scores than its baseline variant, particularly in BLEU4 and METEOR, confirming its ability to generate more semantically aligned reports. However, CE metrics such as Precision, Recall, F1, and RadF1 were not applicable for R2Gen in this setting, as the MIMIC-CXR dataset we used lacks integration with structured clinical annotations.
    
    MLRG demonstrated marked improvements in CE metrics when enhanced with Med-GLIP. These results are attributable to MLRG’s capability to utilize bounding box-derived visual features as part of its multi-view contrastive learning and alignment mechanisms. The model’s ability to encode clinical dependencies and disease progression pathways further enhanced its factual consistency, as reflected in the CE metrics.
    
    % In summary, while R2Gen benefits more from linguistic supervision and excels in NLG-focused metrics due to its transformer architecture, MLRG shows broader utility in clinical settings. It achieves improved clinical fidelity by aligning visual and textual representations through contrastive learning and prior knowledge integration. The performance gained with our dataset demonstrates the feasibility of using bounding box-enhanced data for improving the quality and reliability of automated radiology report generation.

% \vspace{-1em}
\textbf{Performance of downstream Med-VQA tasks.} As shown in \autoref{VQA_Comparison}, Med-GLIP can enhance the performance of downstream Med-VQA tasks. In closed-end tasks, measured by accuracy, Med-GLIP brings gains across all datasets, with the highest improvement of 2.52\% on VQA-RAD. For open-end tasks, evaluated by Rouge-L, it also provides consistent boosts, achieving a maximum uplift of 1.68 on VQA-RAD. These results highlight Med-GLIP's effectiveness in improving Med-VQA performance.

\section*{Conclusion}

% In this work, we introduce \textbf{Med-GLIP}, a unified framework for medical image grounding, alongside a large-scale benchmark, Med-GLIP-5M, which addresses key challenges in the domain—including limited data availability, high modality heterogeneity, lack of fine-grained spatial annotations, and weak generalizability to downstream tasks. 
We present Med-GLIP, a unified and modality-aware framework for medical image grounding, together with Med-GLIP-5M—a large-scale and diverse grounding dataset. Our approach effectively bridges the semantic gap across imaging modalities and significantly improves downstream tasks such as medical VQA and report generation. Extensive experiments validate its superior grounding accuracy and generalization ability. Med-GLIP demonstrates the potential of scalable, spatially grounded pretraining for building generalizable medical vision-language models, paving the way for broader clinical applications and future integration with large language models.
\section*{Acknowledgment}

This work is supported by the National Key R\&D Program of China (Grant No. 2024YFC3308304), the "Pioneer" and "Leading Goose" R\&D Program of Zhejiang (Grant No. 2025C01128), the National Natural Science Foundation of China (Grant No. 62476241), the Natural Science Foundation of Zhejiang Province, China (Grant No. LZ23F020008).

\twocolumn
% 前面是双栏内容

\clearpage
\onecolumn

\clearpage
\twocolumn

\bibliographystyle{IEEEtran}
\bibliography{conference_101719}

% \bibliography{sample-sigconf}

\end{document}